\documentclass[10pt, a4paper]{article}
\usepackage{lrec2022}
\usepackage{multibib}
\newcites{languageresource}{Language Resources}
\usepackage{graphicx}
\usepackage{tabularx}
\usepackage{soul}

\usepackage{subfig}
\usepackage{booktabs}
\usepackage{multirow}
\usepackage{amsmath}
\usepackage{paralist}

\usepackage{titlesec}
\titleformat{\section}{\normalfont\large\bfseries\center}{\thesection.}{1em}{}
\titleformat{\subsection}{\normalfont\SmallTitleFont\bfseries\raggedright}{\thesubsection.}{1em}{}
\titleformat{\subsubsection}{\normalfont\normalsize\bfseries\raggedright}{\thesubsubsection.}{1em}{}
\renewcommand\thesection{\arabic{section}}
\renewcommand\thesubsection{\thesection.\arabic{subsection}}
\renewcommand\thesubsubsection{\thesubsection.\arabic{subsubsection}}

\usepackage{epstopdf}
\usepackage[utf8]{inputenc}
\usepackage{fancyhdr}
\usepackage{hyperref}
\usepackage{xstring}

\usepackage{color}
\usepackage{placeins}

\title{\vspace*{.5\baselineskip}\textbf{An Empirical Study on the Overlapping Problem \\of Open-Domain Dialogue Datasets\\}}

\name{Yuqiao Wen,\ Guoqing Luo,\ Lili Mou}

\address{Dept.~Computing Science, Alberta Machine Intelligent Institute (Amii), University of Alberta\\
yuqiao@ualberta.ca, gluo@ualberta.ca, doublepower.mou@gmail.com\\
}

\abstract{
Open-domain dialogue systems aim to converse with humans through text, and dialogue research has heavily relied on benchmark datasets.
In this work, we observe the overlapping problem in DailyDialog and OpenSubtitles, two popular open-domain dialogue benchmark datasets.
Our systematic analysis then shows that such overlapping can be exploited to obtain fake state-of-the-art performance.
Finally, we address this issue by cleaning these datasets and setting up a proper data processing procedure for future research.
 \\ \newline \Keywords{Open-domain dialogue, Data cleaning}
}

\begin{document}

\maketitleabstract

\section{Introduction}
\renewcommand{\headrulewidth}{0pt}
\cfoot{In LREC'22.}
\thispagestyle{fancy}

Open-domain dialogue generation is the task of generating natural language utterances to converse with humans~\cite{shang2015neural,seq2bf,context}.
Such systems have wide applications in the industry.
For example, XiaoIce\footnote{\url{http://www.xiaoice.com}} from Microsoft has been deployed to more than 40 platforms and has gained 660 million active users since its launch in 2014 \cite{xiaomi}.

There have been several benchmark datasets for open-domain dialogue generation, and they are largely advancing the field.
For example, DailyDialog \cite{li2017dailydialog} and OpenSubtitles \cite{lison2018opensubtitles2018} have been extensively used in recent studies~\cite{cai2020data,sun-etal-2021-generating,zhou-etal-2021-learning,wang-etal-2021-diversifying}.

However, we observe a common problem of existing open-domain dialogue datasets: the training and test sets tend to overlap with each other. That is, a large number of identical or near-identical dialogues appear in both training and test sets, probably due to mistakes in data collection and preprocessing.

We further observe that such overlaps cause bizarre behaviors in training dialogue systems:

\begin{compactenum}
    \item Common evaluation metrics are heavily inflated.
    A dialogue system can achieve perfect test performance on overlapping samples by memorizing the training set. However, such high performance is fake, as it reflects the dialogue system's ability to memorize rather than its conversational skills.
    \item Reported performance based on overlapping samples is arbitrary.
    Due to overlapping, performance may continue to grow even after 1000 epochs as the dialogue system continues to memorize more training samples.
    Since most researchers do not train their models long enough (e.g., for 1000 epochs), their reported performance is arbitrary depending on the maximum epoch or the early stopping strategy, making the comparison of state-of-the-art models meaningless.
    \item Generated utterances are over-informative.
    For example, we observe that models trained on overlapping samples can predict the speaker's name correctly with no context.
    Such behaviour is not realistic for any dialogue systems, highlighting the issues with overlapping datasets.
\end{compactenum}

Therefore, we argue that it is crucial to revisit and clean benchmark dialogue datasets for rigorous scientific research.
Our contributions are threefold.
First, we observe and report the overlap issue of existing dialogue benchmark datasets.
Second, we perform systematic analyses to show the consequences of such overlapping.
Third, we provide cleaned datasets\footnote{Our cleaned datasets and source code are available at: \url{https://github.com/yq-wen/overlapping-datasets}}
for future open-domain dialogue research.

In light of our research, we advocate the following practice in future open-domain dialogue research:
\begin{compactenum}
\item Avoid comparing state-of-the-art models on overlapping datasets; and
\item Always revisit the quality of existing and future datasets for dialogue research.
\end{compactenum}

\section{Related Work}

Dialogue systems can be categorized into two main paradigms: task-oriented and open-domain.
Task-oriented dialogue systems aim to achieve specific tasks such as finding restaurants and booking hotels.
For example, the ATIS corpus~\cite{hemphill1990atis} is an early task-oriented dataset that focuses on air travel, collected through the Wizard-of-Oz~\cite{kelley1984iterative} scheme.
Recently, \newcite{wen2017woz} adopt Wizard-of-Oz to crowd-sourcing and largely reduce the cost for collecting large annotated data.
Following their approach, many more datasets are collected, such as the Frames corpus~\cite{el-asri-etal-2017-frames}, MultiWOZ~\cite{budzianowski2018multiwoz}, and CrossWOZ~\cite{zhu2020crosswoz}.

On the other hand, open-domain dialogue systems aim to hold engaging and open-ended conversations with humans.
Early systems such as ELIZA~\cite{weizenbaum1966eliza} and ALICE~\cite{wallace2009anatomy} are based on manually crafted rules.
Recent advances in deep neural networks make it possible to train dialogue systems end-to-end using massive dialogue corpora~\cite{shang2015neural,li-etal-2016-diversity,serban2016building}.
Therefore, researchers have created many open-domain dialogue datasets, such as DailyDialog~\cite{li2017dailydialog}, Persona-Chat~\cite{zhang2018personalizing}, and Topical-Chat~\cite{gopalakrishnan2019topical}.

While benchmarking datasets largely advance the NLP field over the past decades, it is not uncommon that benchmarked datasets have flaws. For example, the task-oriented dialogue dataset ATIS is adapted to the semantic parsing task. As pointed out by \newcite{guo-etal-2020-benchmarking-meaning} and \newcite{huang-etal-2021-globally}, a large number of samples become identical when researchers anonymize the entities in an utterance~\cite{dong-lapata-2016-language}.
\newcite{schumann2020discrete} identify a problem in the summarization task that the previous benchmark setting does not properly enforce summary length, allowing ``state-of-the-art'' models to gain performance by generating over-lengthed summaries.
These highlight the importance of properly benchmarking a task for NLP research.

The overlapping problem of DailyDialog was first reported by the previous work of one of our coauthors \cite{bahuleyan2019stochastic,coling}. However, the community has not been adequately aware of this. Our new paper studies the problem more systematically. We show that the problem is more severe than we thought as it occurs in different datasets. We perform more detailed analyses and provide an approach to deduplicate overlapping samples.

\section{Bizarre Behaviors When Data Overlap}
\label{sec:bizzare}

We present our empirical findings for the overlapping problem and its consequences with two commonly used open-domain dialogue datasets:
DailyDialog\footnote{\url{http://yanran.li/files/ijcnlp_dailydialog.zip}, accessed on Oct 27, 2021.}
and OpenSubtitles\footnote{Since the original OpenSubtitles dataset does not provide data splits and there is no standard practice, we use the splits from a recent study \cite{wang-etal-2021-diversifying} as a representative:
\url{https://drive.google.com/file/d/1U4M0h9tLNeCyu9JBfSgR3r5EE6IIqyNZ/}, accessed on Nov 8, 2021}.
While overlapping samples exist in many datasets, we find the problem most severe in DailyDialog and OpenSubtitles.

\textbf{Overlapping Statistics. }
We represent an utterance as a bag of words, and compute the overlap ratio between two utterances $\mathbf u=\{u_1,\cdots, u_m\}$ and $\mathbf v=\{v_1,\cdots, v_n\}$ as $R(\mathbf u, \mathbf v)=\frac{2|\mathbf u\cap \mathbf v|}{|\mathbf u| + |\mathbf v|}$. A data sample is a tuple $\mathbf x=(\mathbf c, \mathbf r)$, where $\mathbf c$ is the context and $\mathbf r$ is the response.
We then compute the overlap ratio of two samples $\mathbf x=(\mathbf c, \mathbf r)$ and $\mathbf x'=(\mathbf c', \mathbf r')$ as $R(\mathbf x, \mathbf x')=\min\{R(\mathbf c, \mathbf c'), R(\mathbf r, \mathbf r')\}$; the $\min$ operator rules out false overlapping caused by generic utterances (e.g., \textit{hello}), whose responses may be different.
Finally, the overlap ratio of a test sample $\mathbf x$ against the training dataset $\mathcal D_\text{train}$ is given by $R(\mathbf x, \mathcal D_\text{train})=\max_{\mathbf x'\in\mathcal D_\text{train}} R(\mathbf x, \mathbf x')$.

\begin{figure}[!t]%
    \centering
    \subfloat
        {{\includegraphics[width=0.25\textwidth]{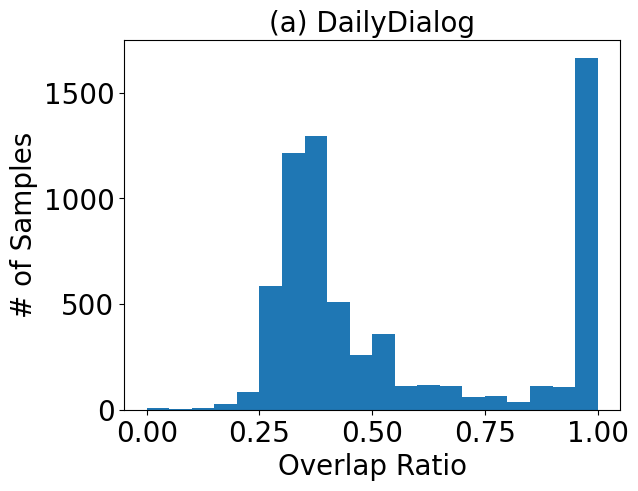}}}
    \subfloat
        {{\includegraphics[width=0.25\textwidth]{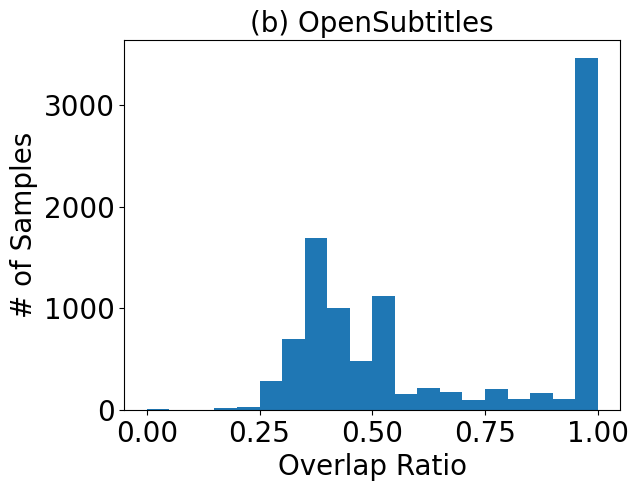}}}
    \caption{Overlap histogram of test samples against the training set on (a) DailyDialog and (b) OpenSubtitles}%
    \label{fig:cumulative}%
\end{figure}

Figure~\ref{fig:cumulative} shows the histogram of the overlap ratio in DailyDialog and OpenSubtitles.
We find that 23.15\% and 34.49\% test samples are identical to training samples in the two datasets, respectively.
Further, many samples are near-identical.
In Table~\ref{tab:overlap}, for instance, a test sample only contains additional speaker information (``A~::'' and ``B~::'') compared with another training sample, resulting in an overlap ratio of 0.80.

As shown, such overlapping does not naturally arise from human conversations;
rather, they are caused by oversights in data collection and preprocessing.
For DailyDialog, the dataset is constructed by crawling English learning websites \cite{li2017dailydialog}, and overlapping samples probably come from the common learning materials shared among different websites.
For OpenSubtitles, \newcite{lison2018opensubtitles2018} group subtitles based on their IMDb identifiers.
However, we find that the same movie may correspond to different identifiers if it has different versions.
For example, the South Korean movie My Sassy Girl\footnote{\url{https://www.imdb.com/title/tt0293715/}} and its American remake\footnote{\url{https://www.imdb.com/title/tt0404254/}} contain highly overlapping dialogues, but they are treated as different movies based on their IMDb identifiers.
Such overlapping raises serious concerns on whether these datasets are appropriate for benchmarking open-domain dialogue research.

\begin{table}[!t]
\resizebox{\linewidth}{!}{%
\begin{tabular}{l|l|l|l}
\hline
\multirow{4}{*}{0.60} & \multirow{2}{*}{Train} & Context & Do you have a fever ?                 \\
 &                       & Response & I don't know , but I feel terrible .         \\ \cline{2-4}
 & \multirow{2}{*}{Test} & Context  & Do you have an airsickness ?                 \\
 &                       & Response & I don't know . But I have a carsickness .    \\ \hline
\multirow{4}{*}{0.80} & \multirow{2}{*}{Train} & Context & Nice to meet you , Mr . Wilson .      \\
 &                       & Response & Tim , please . Please be seated .            \\ \cline{2-4}
                      & \multirow{2}{*}{Test}  & Context  & B :: Nice to meet you , Mr . Wilson . \\
 &                       & Response & A :: Tim , please . Please be seated .       \\ \hline
\multirow{4}{*}{1.00} & \multirow{2}{*}{Train} & Context & It seldom rains this summer .         \\
 &                       & Response & Yeah , some places are very short of water . \\ \cline{2-4}
 & \multirow{2}{*}{Test} & Context  & It seldom rains this summer .                \\
 &                       & Response & Yeah , some places are very short of water . \\ \hline
\end{tabular}%
}
\caption{Training and test samples with their corresponding overlap ratios from the original DailyDialog dataset.
}
\label{tab:overlap}
\end{table}

\textbf{Learning Curves.}
We observe that overlapping samples have an undesired effect on training dialogue systems.
We compare the learning curves on the original test set and a deduplicated test set, where
we remove samples with an overlap ratio of greater than 0.80.
In terms of the neural architecture, we fine-tune a T5-small model~\cite{2020t5} here.

\begin{figure}[!t]%
    \centering
    \subfloat
        {{\includegraphics[width=0.25\textwidth]{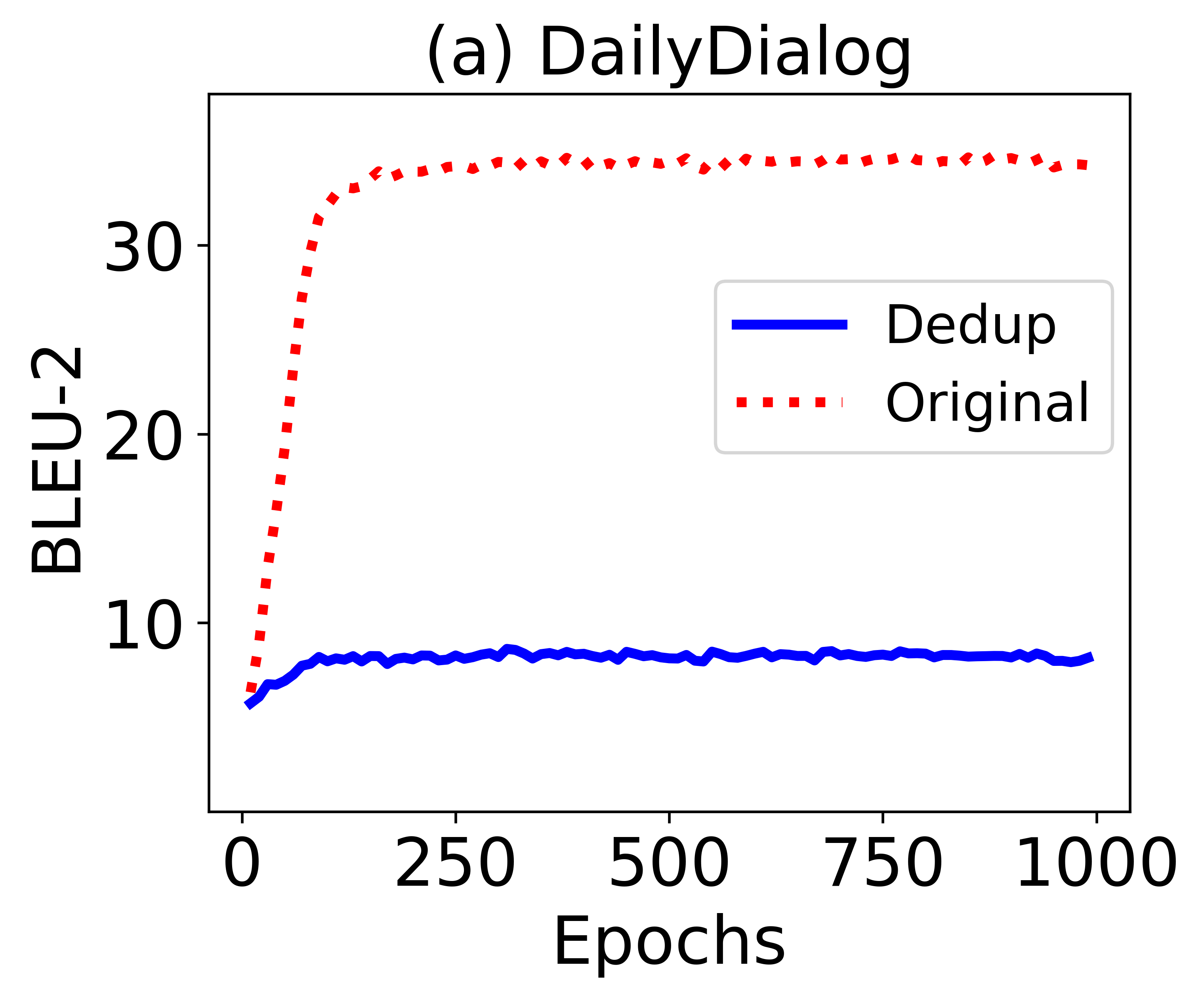}}}
    \subfloat
        {{\includegraphics[width=0.25\textwidth]{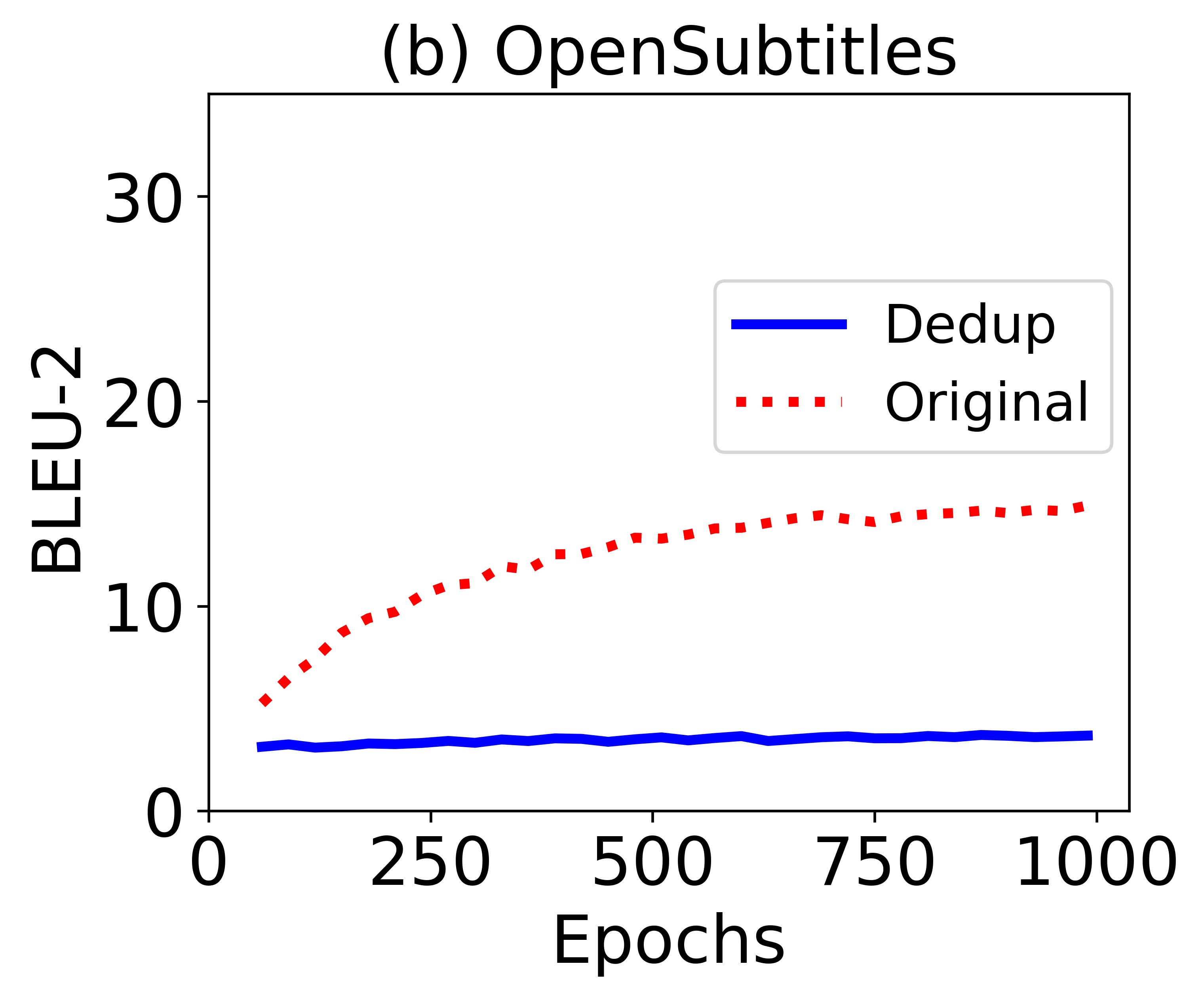}}}
    \caption{
    BLEU-2 learning curves for the original dataset and the deduplicated dataset for (a) DailyDialog and (b) OpenSubtitles.
    Samples with an overlap greater than 0.80 are considered duplicates and are removed for the deduplicated dataset.
    }%
    \label{fig:lc}%
\end{figure}

Figure~\ref{fig:lc} shows the learning curves in terms of the BLEU-2 metric.
On DailyDialog, the model achieves $\sim$34 BLEU-2 with the original test set; however, the same model only achieves $\sim$8 BLEU-2 after deduplication. The trend is similar on OpenSubtitles: 15 \mbox{BLEU-2} with the original test set versus 3 BLEU-2 with the deduplicated one. The results suggest that overlapping samples heavily inflate BLEU scores, and that the high performances of the alleged ``state-of-the-art'' models mainly come from memorizing training samples.

In addition, we observe that it takes more epochs for the test set performance to converge when the test samples overlap with the training set.
For example, the \mbox{BLEU-2} score still increases after 1000 epochs on OpenSubtitles, which contains more complex and diverse conversations than the DailyDialog dataset.
Since most researchers do not train their models for 1000 epochs, their reported performance may be arbitrary, depending on where the model ends up along the learning curve: \newcite{wang-etal-2021-diversifying} report 9.8 BLEU-2 on OpenSubtitles, whereas \newcite{sun-etal-2021-generating} report 32.6.
This highlights the inconsistency of reported performances.

\textbf{Overly Informative Outputs.}
Table~\ref{tab:overinformative} illustrates another bizarre behavior that the model generates overly informative outputs. For example, we consider fine-tuning T5-small for single-turn conversations. Given the input \textit{Nice to see you, Patrick}, the model generates \textit{Bob! I hear your team won the match}.
The output precisely matches the test reference, achieving a BLEU score of 100. Such an output is overly informative because it is extremely unlikely that the model can correctly predict the speaker's name without any context.

In summary, our qualitative and quantitative analyses show that it is fundamentally flawed to evaluate open-domain dialogue systems on overlapping datasets, which are unfortunately commonly used (and benchmarked) in current research.

\begin{table}[!t]
\centering
\resizebox{\linewidth}{!}{%
    \begin{tabular}{|l|l|}
    \hline
    \multicolumn{2}{|c|}{\textbf{DailyDialog}}\\ \hline
    Train Input &  Nice to see you , Patrick .              \\
    Train Ref &  Bob !  I hear your team won the match .    \\ \hline
    Test Input & Nice to see you , Patrick .               \\
    Test Ref &  Bob ! I hear your team won the match .      \\
    Model Output &  Bob !  I hear your team won the match . \\ \hline\hline
    \multicolumn{2}{|c|}{\textbf{OpenSubtitles}}\\ \hline
    Train Input &  But you have some strength in you , my dear Hobbit .          \\
    Train Ref &  What happened, Gandalf ?              \\ \hline
    Test Input & But you have some strength in you , my dear Hobbit .\\
    Test Ref & What happened, Gandalf ?                     \\
    Model Output & What happened, Gandalf ?                                         \\ \hline
    \end{tabular}%
}
\caption{Over-informative model outputs with DailyDialog and OpenSubtitles samples.}
\label{tab:overinformative}
\end{table}

\section{Dataset Cleaning}
\label{sec:data-cleaning}

We make efforts to clean existing datasets and present results for standard and state-of-the-art models on the cleaned corpora.

In the literature, both single-turn and multi-turn settings are common~\cite{cai2020data,zhou-etal-2021-learning,wang-etal-2021-diversifying,gu2021dialogbert}. We propose to deduplicate data samples in the multi-turn setting, i.e., a dialogue session in DailyDialog and an entire movie in OpenSubtitles are treated as a unit for deduplication. Such a unit can be further split into single turns, so our treatment unifies both settings.

Further, we propose to re-split the training, validation, and test sets of the original corpora during deduplication. If we simply remove the duplicate samples as in Section~\ref{sec:bizzare}, then at least one of the validation and test sets will be heavily shrunk due to massive overlapping samples, making evaluation noisy and unreliable.

Let us consider two units $\mathbf u$ and $\mathbf v$ (dialogue sessions or movies) for deduplication.
Their overlap ratio  is defined as
\begin{equation}
    \label{eq:ss-overlap}
    R{(\mathbf u, \mathbf v)} = \frac{2|\mathbf u \cap \mathbf v|}{|\mathbf u| + |\mathbf v|}
\end{equation}

The overlap ratio of a unit $\mathbf{u}$ is then computed as the maximum overlap against all other units in the dataset $\mathcal D$:
\begin{equation}
    \label{eq:sd-overlap}
    R(\mathbf{u}, \mathcal D) = \max_{\mathbf{u}' \in \mathcal D \setminus \{\mathbf u\}} R(\mathbf{u}', \mathbf{u})
\end{equation}

\begin{table*}[!t]
\centering
\scalebox{0.9}{
\begin{tabular}{l|l|l|r|r|r}
\hline
Dataset & Version & Context History & \multicolumn{1}{r|}{\# Train} & \multicolumn{1}{r|}{\# Validation} & \multicolumn{1}{r}{\# Test} \\ \hline
\multirow{4}{*}{DailyDialog}   & \multirow{2}{*}{\newcite{li2017dailydialog}}  & Single-Turn & 76,052    & 7,069  & 6,740  \\ \cline{3-6} 
                               &                          & Multi-Turn  & 76,052    & 7,069  & 6,740  \\ \cline{2-6} 
                               & \multirow{2}{*}{Cleaned} & Single-Turn & 60,005    & 6,594  & 6,955  \\ \cline{3-6} 
                               &                          & Multi-Turn  & 60,138    & 6,612  & 6,980  \\ \hline
\multirow{3}{*}{OpenSubtitles} & \newcite{wang-etal-2021-diversifying}                & Single-Turn & 1,144,949 & 20,000 & 10,000 \\ \cline{2-6} 
                               & \multirow{2}{*}{Cleaned} & Single-Turn & 979,230   & 11,982 & 12,152 \\ \cline{3-6} 
                               &                          & Multi-Turn  & 1,002,026 & 12,289 & 12,506 \\ \hline
\end{tabular}%
}
\vspace{-2mm}
\caption{
    Data statistics for DailyDialog and OpenSubtitles.
}
\vspace{-2mm}
\label{tab:data-stats}
\end{table*}

Note that the above overlap ratio is similar to the one in Section~\ref{sec:bizzare} (Overlapping Statistics).
However, Eqn.~(\ref{eq:sd-overlap}) considers the ratio of a deduplication unit (an entire dialogue session or movie), whereas Section~\ref{sec:bizzare} considers the ratio of a single-turn conversation. Thus, we do not have the $\operatorname{min}$ operator here. Further, the ratio of a unit is computed against the rest of the corpus, whereas Section~\ref{sec:bizzare} computes the ratio of a test sample against the training set. Thus, we have the set minus operator in Eqn.~(\ref{eq:sd-overlap}).

Then, we iterate through all deduplication units in the corpus.
If a unit's overlap ratio exceeds a threshold, then we remove the unit but keep the one that it overlaps the most with.
Since there may be more than two units overlapping with each other, we need to repeat the procedure multiple times.
Specifically, we recompute the overlap ratios after each iteration through the dataset, and remove additional duplicate samples until convergence.

After that, we split the deduplicated units into training, validation, and test sets.
While our deduplication unit is an entire dialogue session or a movie, the resulting corpus can serve for the settings of single-turn and fixed-turn context. We simply split a unit into multiple tuples of contexts and responses.
This may result in (a small number of) additional duplicate samples due to the generic conversations, such as \textit{--Thank you. --You're welcome}. Therefore, we further remove exactly overlapping context--response pairs.

Table \ref{tab:data-stats} shows the data statistics for our cleaned datasets for both single-turn and multi-turn settings.
Specifically, we follow previous work and use three turns for the multi-turn setting~\cite{li2017dailydialog,cai2020learning,wang-etal-2021-diversifying}.
Our data cleaning strategy enables us to control the size of validation and test sets.
For DailyDialog, we keep the size to be similar to the original dataset, which cannot be achieved by a na\"ive removal of validation/test samples.
For OpenSubtitles, the corpus contains much more samples than needed for training a dialogue system; therefore, we follow previous work~\cite{du2018variational,cai2020data,wang-etal-2021-diversifying} and sample a similar number of  training, validation, and test samples.

\section{Model Performance}

\begin{table*}[t!]
\centering
\resizebox{1.00\linewidth}{!}{
\begin{tabular}{c|l|rrrr|rrrr}
\hline
\multirow{3}{*}{Context History} &
  \multicolumn{1}{c|}{\multirow{3}{*}{Model}} &
  \multicolumn{4}{c|}{\multirow{2}{*}{Cleaned DailyDialog}} &
  \multicolumn{4}{c}{\multirow{2}{*}{Cleaned OpenSubtitles}} \\
                             & \multicolumn{1}{c|}{} & \multicolumn{4}{c|}{}                                  & \multicolumn{4}{c}{}                                   \\ \cline{3-10} 
                             & \multicolumn{1}{c|}{} & BLEU-2 & \multicolumn{1}{r|}{BLEU-4} & Dist-1 & Dist-2 & BLEU-2 & \multicolumn{1}{r|}{BLEU-4} & Dist-1 & Dist-2 \\ \hline
\multirow{6}{*}{Single-Turn} & LSTM w/ attn          & 6.56   & \multicolumn{1}{r|}{2.11}   & 3.40   & 23.50  & 5.31   & \multicolumn{1}{r|}{1.41}   & 3.10   & 14.94  \\
                             & Transformer           & 7.33   & \multicolumn{1}{r|}{2.56}   & 4.16   & 25.44  & 4.89   & \multicolumn{1}{r|}{1.29}   & 3.05   & 13.88  \\
                             & T5-small              & 8.74   & \multicolumn{1}{r|}{3.39}   & 4.63   & 25.43  & 6.76   & \multicolumn{1}{r|}{2.07}   & 2.78   & 8.87   \\
                             & GPT-2                 & 7.67   & \multicolumn{1}{r|}{2.78}   & 5.38   & 29.15  & 7.02   & \multicolumn{1}{r|}{2.15}   & 2.98   & 11.37  \\
                             & AdaLabel              & 6.72   & \multicolumn{1}{r|}{2.29}   & 4.35   & 26.97  & 5.66   & \multicolumn{1}{r|}{1.45}   & 3.86   & 15.33  \\
                             & DialogBERT$^\dagger$            & 5.42   & \multicolumn{1}{r|}{2.16}   & 2.57   & 19.53  & 3.29   & \multicolumn{1}{r|}{0.46}   & 2.62   & 19.38  \\ \hline
\multirow{6}{*}{Multi-Turn}  & LSTM w/ attn          & 7.06   & \multicolumn{1}{r|}{2.34}   & 3.18   & 22.76  & 4.74   & \multicolumn{1}{r|}{1.10}   & 3.36   & 19.63  \\
                             & Transformer           & 7.35   & \multicolumn{1}{r|}{2.65}   & 4.06   & 25.91  & 4.64   & \multicolumn{1}{r|}{1.21}   & 3.53   & 16.75  \\
                             & T5-small              & 9.49   & \multicolumn{1}{r|}{3.81}   & 4.77   & 25.83  & 7.38   & \multicolumn{1}{r|}{2.42}   & 2.81   & 9.77   \\
                             & GPT-2                 & 8.55   & \multicolumn{1}{r|}{3.39}   & 5.12   & 27.75  & 7.26   & \multicolumn{1}{r|}{2.28}   & 3.13   & 12.24  \\
                             & AdaLabel              & 6.13   & \multicolumn{1}{r|}{2.11}   & 4.63   & 28.65  & 5.75   & \multicolumn{1}{r|}{1.41}   & 3.71   & 14.77  \\
                             & DialogBERT$^\dagger$            & 6.34   & \multicolumn{1}{r|}{1.88}   & 5.21   & 30.61  & 3.90   & \multicolumn{1}{r|}{0.68}   & 3.03   & 22.01  \\ \hline
\end{tabular}%
}
\caption{
    Performance of various models on DailyDialog and OpenSubtitles.
    $^\dagger$DialogBERT uses sampling-based decoding following the original implementation. Other models use greedy decoding; we observe DialogBERT failed to perform reasonably with greedy decoding.
}
\label{tab:main-results}
\end{table*}

In this section, we test multiple standard and state-of-the-art models on our cleaned datasets.
We will first show the metrics and models. Then, we will present the experimental results.

\subsection{Metrics}

We use the BLEU score \cite{papineni2002bleu} as an automatic metric to evaluate the quality of the generated responses, following most previous work \cite{cai2020data,sun-etal-2021-generating,wang-etal-2021-diversifying}.
The BLEU metric measures the lexical overlap between the model output and the reference. Note that our \mbox{BLEU-$n$} measures the geometric mean of $i$-gram overlap for $i=1,\cdots,n$.

Additionally, we adopt the Dist metric \cite{li-etal-2016-diversity} to measure the diversity of the generated responses, which is another commonly used metric in dialogue research \cite{xu-etal-2017-neural,zhou-etal-2021-learning,wang-etal-2021-diversifying}.
Specifically, Dist-$n$ measures the percentage of distinct $n$-grams among all generated responses.

\subsection{Models}

We evaluate multiple standard and state-of-the-art models on our proposed datasets.

\begin{compactitem}
\item \textbf{LSTM with Attention.} We include the long short-term memory network \cite{hochreiter1997long} with an attention mechanism \cite{bahdanau2015neural} as a baseline. This is a standard model before the Transformer era.

\item\textbf{Transformer.} In current research, the Transformer \cite{vaswani2017attention} is the most commonly used architecture. It replaces LSTM's recurrent connections with a multi-head attention mechanism and achieves superior performance in different NLP tasks. In this baseline, the Transformer is not pretrained.

\item\textbf{GPT-2.} The GPT-2 model~\cite{gpt2} adopts the Transformer architecture, but is pretrained on massive unlabeled corpora. Pretraining is shown to benefit various downstream tasks.

\item\textbf{T5-small.} The T5 model~\cite{2020t5} also uses the Transformer architecture, but works in an encoder--decoder fashion. It is pretrained on a number of text generation tasks, such as translation and summarization. We adopt the T5-small version in our experiments.

\item\textbf{AdaLabel.} Being one of the recent state-of-the-art models, AdaLabel~\cite{wang-etal-2021-diversifying} also uses the  Transformer encoder--decoder architecture. Instead of the standard cross-entropy training, the model is trained with the adaptive label smoothing technique, where the one-hot labels are smoothed by soft predictions generated from an auxiliary decoder.

\item\textbf{DialogBERT.} DialogBERT~\cite{gu2021dialogbert} is another recent state-of-the-art model. Instead of using a standard Transformer encoder, it uses a hierarchical BERT~\cite{devlin2019bert} encoder trained with masked utterance regression and distributed utterance order ranking, so as to better capture discourse-level dialogue context.
\end{compactitem}

In our experiments, we specifically choose AdaLabel and DialogBERT because they are recently published models from top-tier venues using overlapping datasets. Moreover, the authors publish their code\footnote{AdaLabel: \url{https://github.com/lemon234071/AdaLabel}; DialogBERT: \url{https://github.com/guxd/DialogBERT}},  allowing us to replicate their work on our cleaned datasets.

\textbf{Setup.}
We use the fairseq~\cite{ott2019fairseq} framework to train the LSTM model and the Transformer model.
The LSTM model has a single layer with a hidden size of 512 for both the encoder and decoder.
The Transformer model has 6 layers with a hidden size of 512 with 8 attention heads.
Our pretrained models are adopted from HuggingFace~\cite{wolf2019huggingface} and fine-tuned in the experiments.
Specifically, T5-small uses 6 layers with a hidden size of 512 and 8 attention heads, whereas GPT-2 uses 12 layers with a hidden size of 768 and 12 attention heads.
For AdaLabel and DialogBERT, we directly run their source code on our cleaned datasets.
For AdaLabel, both the encoder and decoder contain 6 layers with a hidden size of 512 and 8 attention heads.
DialogBERT uses 6 transformer layers, a hidden size of 256, and 2 attention heads for the utterance encoder, context encoder, and decoder.
All models are trained with the Adam optimizer~\cite{DBLP:journals/corr/KingmaB14}.

\subsection{Results}

Table \ref{tab:main-results} shows the performance of baseline and state-of-the-art models on our cleaned datasets.
For completeness, we evaluate all models on both single-turn and multi-turn settings, which is common in previous studies~\cite{sun-etal-2021-generating,cai2020learning,wang-etal-2021-diversifying}.

As seen, the simple LSTM model is on par with the Transformer model: it is slightly worse on DailyDialog but better on OpenSubtitles.
The pretrained models outperform the un-pretrained ones, which is consistent with existing  literature such as machine translation \cite{lewis-etal-2020-bart,liu-etal-2020-multilingual-denoising,2020t5}.

However, the performance gap between different baseline models is noticeably smaller in other text generation tasks.
For example, \newcite{luong2015effective} achieve a BLEU score of 20.9 on the WMT'14 English--German dataset using LSTM with attention, whereas \newcite{vaswani2017attention} achieve 27.3 with a Transformer model, giving an improvement of 6.4 points.
\newcite{2020t5} further improve the model by 3.6 points with pretraining techniques, achieving a BLEU score of 30.9.
We hypothesize that this is due to the inherent uncertainty of the dialogue task, which may not be fully alleviated by pretraining techniques.

We then compare the above baselines with several alleged state-of-the-art models: AdaLabel \cite{wang-etal-2021-diversifying} and DialogBERT~\cite{gu2021dialogbert}, because their models are open-sourced.

AdaLabel is a Transformer-based encoder--decoder model, enhanced with an adaptive label smoothing technique.
Based on the cleaned datasets, we find that AdaLabel's performance is (at most) on par with the pretrained models. On the one hand, AdaLabel yields consistently lower BLEU scores than GPT-2. On the other hand, AdaLabel appears to achieve higher diversity scores in some settings, but is much lower in others (e.g., single-turn DailyDialog).

Another alleged state-of-the-art model is DialogBERT, which uses hierarchical BERT to encode multi-turn context.
In our experiments, we were unable to obtain reasonable performance with greedy decoding since the model soon converges to generic responses.
Therefore, we keep the the same sampling setting as in the original paper, which leads to high diversity scores (given by the Dist metric).
However, a higher diversity score does not necessarily imply a better dialogue system, as a random decoder will achieve the highest Dist scores. In fact, DialogBERT results in very low BLEU scores (the main metric in most literature): it is worse than all models in all settings, except the LSTM in single-turn DailyDialog.

Overall, it is highly questionable whether these alleged state-of-the-art models truly outperform standard baselines.  Our observations contradict the original papers' experiments \cite{wang-etal-2021-diversifying,gu2021dialogbert}, where the authors use overlapping datasets and show the superior performance of the proposed models in terms of both BLEU and Dist.
Such a discrepancy is largely due to the overlapping samples: although their proposed models achieve a lower BLEU score in terms of conversational skills, they can boost it by memorizing overlapping samples, showing unreal improvement in both BLEU and Dist.
This confirms the importance of cleaning existing benchmark datasets for future dialogue research.

\section{Conclusion}

In this work, we observe the overlapping issues in two popular open-domain dialogue benchmark datasets: DailyDialog and OpenSubtitles.
We then conduct systematic analyses and show the undesired consequences when data overlap.
To address the issue, we propose a data cleaning strategy to set up a proper protocol for future research.
Our experiments on the cleaned datasets show that the ``state-of-the-art'' performance on overlapping datasets is questionable, highlighting the importance of revisiting open-domain dialogue datasets.

\section{Acknowledgments}
The research is supported in part by the Natural Sciences and Engineering Research Council of Canada (NSERC) under grant
No.~RGPIN2020-04465, the Amii Fellow Program, the Canada CIFAR AI Chair Program, a UAHJIC project, a donation from DeepMind, and Compute Canada (www.computecanada.ca).

\FloatBarrier

\section{Bibliographical References}\label{reference}

\bibliographystyle{lrec2022-bib}
\bibliography{lrec2022-example}

\end{document}